\documentclass[a4paper,11pt]{article}
\usepackage{pos}

\title{Machine learning with quantum field theories}

\author*[a]{Dimitrios Bachtis}
\author[b,c]{Gert Aarts}
\author[a,d]{Biagio Lucini}

\affiliation[a]{Swansea University,\\
  Department of Mathematics, Bay Campus, SA1 8EN, Swansea, Wales, United Kingdom}

\affiliation[b]{ Swansea University,\\ Department of Physics, Singleton Campus, SA2 8PP, Swansea, Wales, United Kingdom}
\affiliation[c]{European Centre for Theoretical Studies in Nuclear Physics and Related Areas (ECT*) \\ \& Fondazione Bruno Kessler
Strada delle Tabarelle 286, 38123 Villazzano (TN), Italy }

\affiliation[d]{Swansea University,\\
Swansea Academy of Advanced Computing, Bay Campus, SA1 8EN, Swansea, Wales, United Kingdom}
\emailAdd{dimitrios.bachtis@swansea.ac.uk}
\emailAdd{g.aarts@swansea.ac.uk}
\emailAdd{b.lucini@swansea.ac.uk}

\abstract{The precise equivalence between discretized Euclidean field theories and a certain class of probabilistic graphical models, namely the mathematical framework of Markov random fields, opens up the opportunity to investigate machine learning from the perspective of quantum field theory. In this contribution we will demonstrate, through the Hammersley-Clifford theorem, that the $\phi^{4}$ scalar field theory on a square lattice satisfies the local Markov property and can therefore be recast as a Markov random field. We will then derive from the $\phi^{4}$ theory machine learning algorithms and neural networks which can be viewed as generalizations of conventional neural network architectures. Finally, we will conclude by presenting applications based on the minimization of an asymmetric distance between the probability distribution of the $\phi^{4}$ machine learning algorithms and target probability distributions.}

\FullConference{%
 The 38th International Symposium on Lattice Field Theory, LATTICE2021
  26th-30th July, 2021
  Zoom/Gather@Massachusetts Institute of Technology
}

\newtheorem{thm}{Theorem}
\usepackage{bm}
\begin{document}
\maketitle

\section{Introduction}
The transition to Euclidean space and the discretization of quantum field theories on a graph representation are two fundamental steps undertaken to formulate quantum fields within the framework of lattice field theory. Lattice field theory, which shares inherent connections with statistical physics and probability theory, provides a rigorous approach to define mathematically and  to simulate numerically quantum field theories. It is through the probabilistic perspective that a connection between lattice field theories and a certain class of machine learning algorithms, namely the mathematical framework of Markov random fields, can be rigorously established~\citep{bachtis-qftml}.

A Markov random field can be defined as a set of random variables on a graph structure that satisfy Markov properties. Markov properties are significant conditions of locality that emerge across different research fields. Within the constructive aspect of quantum field theory, Markov properties appear as required conditions to rigorously construct quantum fields in Minkowski space based on Markov fields in Euclidean space~\citep{nelson-qft}. In machine learning~\citep{fischer-pgm}, algorithms that satisfy Markov properties have been used extensively in problems  ranging from computer vision to computational biology, where the imposed condition of locality is exploited to uncover local structures in data.

In this contribution, we derive machine learning algorithms and neural networks from lattice field theories, opening up the opportunity to investigate machine learning directly within quantum field theory. Specifically, we demonstrate, through the Hammersley-Clifford theorem, that the $\phi^{4}$ scalar field theory on a square lattice satisfies the local Markov property with respect to the graph structure, therefore recasting it as a Markov random field. We then derive neural network architectures from the $\phi^{4}$ theory which can be viewed as generalizations of conventional architectures. Finally, we present applications pertinent to the minimization of an asymmetric distance between the probability distribution of the $\phi^{4}$ machine learning algorithms and target probability distributions.

\section{The $\phi^{4}$ theory as a Markov random field}

\begin{figure}[b]
\center
\includegraphics[width=6.5cm]{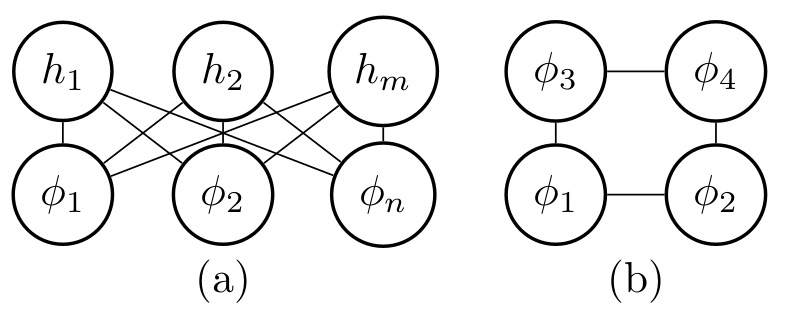}
\caption{\label{fig:graph} A bipartite graph (a) and a square lattice (b). Examples of maximal cliques are $\lbrace \phi_{1}, h_{1} \rbrace$ and $\lbrace \phi_{3}, \phi_{4} \rbrace$, respectively. }
\end{figure}

We consider a physical model which is represented by a finite set $\Lambda$ whose points lie on the vertices of a finite graph $\mathcal{G}=(\Lambda,e)$, where $e$ is the set of edges in $\mathcal{G}$. Two points $i,j \in \Lambda$ are called neighbours if there exists an edge between $i$ and $j$. A clique is a subset of $\Lambda$ whose points are neighbours, and a clique is called maximal if no additional point can be included such that the resulting set is still a clique, see Fig.~\ref{fig:graph}. We now associate to each point $i \in \Lambda$ a random variable $\phi_{i}$.

Given a graph $\mathcal{G}=(\Lambda,e)$ the random variables define a Markov random field if the associated probability distribution $p(\phi)$ satisfies the local Markov property with respect to the graph structure $\mathcal{G}$:
\begin{equation}
p(\phi_{i} | (\phi_{j})_{j \in \Lambda}) = p(\phi_{i} | (\phi_{j})_{j \in \mathcal{N}_{i}}),
\end{equation}
where $\mathcal{N}_{i}$ is the set of neighbours of a given point $i$. We will prove the local Markov property for a set of random variables using the Hammersley-Clifford theorem.

\begin{thm}[Hammersley-Clifford]
A strictly positive distribution $p$ satisfies the local Markov property of an undirected graph $\mathcal{G}$, if and only if $p$ can be represented as a product of strictly positive potential functions $\psi_{c}$ over $\mathcal{G}$, one per maximal clique $c \in C$, i.e.,
\begin{equation}
p(\phi)= \frac{1}{Z} \prod_{c \in C} \psi_{c}(\phi),
\end{equation}
where $Z=\int_{\bm{\phi}} \prod_{c \in C} \psi_{c}(\bm{\phi}) d\bm{\phi}$ is the partition function and $\bm{\phi}$ are all possible states of the system.
\end{thm}

We now consider the two-dimensional $\phi^{4}$ theory on a square lattice with the Euclidean action
\begin{equation}\label{eq:midaction}
S_{E}= -\kappa_{L}\sum_{\langle ij \rangle}\phi_{i} \phi_{j} + \frac{(\mu_{L}^{2}+4\kappa_{L})}{2} \sum_{i} \phi_{i}^{2} +  \frac{\lambda_{L}}{4}  \sum_{i}\phi_{i}^{4}.
\end{equation}
The dimensionless parameters $\kappa_{L},\mu_{L}^{2},\lambda_{L}$ are now considered as inhomogeneous and are redefined as $w=\kappa_{L}$, $a=(\mu_{L}^{2}+4\kappa_{L})/2$, $b=\lambda_{L}/4$, giving rise to the action
\begin{equation}\label{eq:finalaction}
S(\phi ; \theta)= -\sum_{\langle ij \rangle} w_{ij} \phi_{i}\phi_{j} + \sum_{i} a_{i} \phi_{i}^{2} + \sum_{i} b_{i} \phi_{i}^{4},
\end{equation}
where the set of coupling constants is $\theta=\lbrace w_{ij}, a_{i},b_{i} \rbrace$, and the associated Boltzmann probability distribution is:
\begin{equation} \label{eq:probmrf}
p(\phi ; \theta)=\frac{\exp\big[-S(\phi ; \theta)\big]}{\int_{\bm{\phi}}{\exp[-S(\bm{\phi},\theta)]} d\bm{\phi}}.
\end{equation}
The $\phi^{4}$ lattice field theory is, by definition, formulated on a graph $\mathcal{G}=(\Lambda,e)$. A nonunique choice of a potential function $\psi_{c}$ which factorizes the probability distribution  in terms of maximal cliques $c \in C$ is
\begin{equation}
\psi_{c} = \exp\bigg[ -w_{ij} \phi_{i} \phi_{j}+ \frac{1}{4} (a_{i} \phi_{i}^{2} +a_{j}\phi_{j}^{2}  +b_{i} \phi_{i}^{4} +b_{j}\phi_{j}^{4})\bigg],
\end{equation}
where $i,j$ are nearest neighbours.

By demonstrating that the strictly positive distribution $p(\phi;\theta)$ of the $\phi^{4}$ scalar field theory on a square lattice can be represented as a product of strictly positive potential functions $\psi_{c}$ over the maximal cliques of the graph $\mathcal{G}$, we have verified that the $\phi^{4}$ theory is a Markov random field. 

\section{Machine learning with $\phi^{4}$ Markov random fields}

\subsection{Learning without predefined data}

Within the probabilistic perspective a wide class of machine learning applications can be expressed as the minimization of a distance metric between two probability distributions. Specifically, we are searching for the optimal values of the variational parameters $\theta$, which are the coupling constants of the $\phi^{4}$ theory, that are able to minimize the considered distance metric.

As a metric we consider the Kullback-Leibler divergence, which can be viewed as an asymmetric distance between the probability distribution $p(\phi; \theta)$ of the $\phi^{4}$ machine learning algorithms and a target probability distribution $q(\phi)$:
\begin{equation} \label{eq:kl}
KL(p || q) = \int_{-\infty}^{\infty} {p(\bm{\phi}; \theta)} \ln \frac{p(\bm{\phi}; \theta)}{q(\bm{\phi})} d\bm{\phi}  \geq 0.
\end{equation}

The Kullback-Leibler divergence is a nonnegative quantity which becomes zero if and only if the two probability distributions $p(\phi; \theta)$ and $q(\phi)$ are equal. Our aim is to minimize this quantity through a variational approach, specifically a gradient-based algorithm. After a successful minimization, the two probability distributions will be equal and we can use the probability distribution $p(\phi;\theta)$ to draw samples from the target distribution $q(\phi)$.

We consider a target Boltzmann probability distribution $q(\phi)= \exp [-\mathcal{A}]/Z_{\mathcal{A}}$ which describes an arbitrary statistical system or an arbitrary lattice field theory. By substituting the probability distributions $p(\phi)$ and $q(\phi)$ in the Kullback-Leibler divergence we obtain:
\begin{equation}\label{eq:fen2}
F_{\mathcal{A}} \leq \langle \mathcal{A} - S \rangle_{p(\phi;\theta)} + F \equiv \mathcal{F},
\end{equation}
where $\mathcal{F}$ is the variational free energy and $\langle O \rangle_{p(\phi;\theta)}$ denotes the expectation value of an observable $O$ under the probability distribution $p(\phi;\theta)$. There are two important observations for Eq.~\ref{eq:fen2}: first, it sets a rigorous upper bound to the calculation of the free energy $F_{\mathcal{A}}$ of the target system with action $\mathcal{A}$ and, second, the minimization of this quantity is entirely dependent on calculations conducted under the probability distribution $p(\phi;\theta)$ of the $\phi^{4}$ theory. The implication of these observations is that the system with target action $\mathcal{A}$ can be approximated and, subsequently, sampled using exclusively configurations drawn from the probability distribution $p(\phi;\theta)$ of the $\phi^{4}$ theory.

To minimize the variational free energy $\mathcal{F}$ we implement a gradient-based approach
\begin{equation}
\frac{\partial \mathcal{F}}{\partial \theta_{i}}= \langle \mathcal{A} \rangle \Big\langle \frac{\partial S}{\partial \theta_{i}} \Big\rangle -\Big\langle \mathcal{A} \frac{\partial S}{\partial \theta_{i}} \Big\rangle + \Big\langle S \frac{\partial S}{\partial \theta_{i}} \Big\rangle - \langle S \rangle \Big\langle \frac{\partial S}{\partial \theta_{i}} \Big\rangle,
\end{equation} 
and the variational parameters $\theta$ are then updated at each epoch $t$ based on the relation:
\begin{equation}\label{eq:gas}
\theta^{(t+1)}=\theta^{(t)}-\eta*  \mathcal{L},
\end{equation}
where $\eta$ is the learning rate and $\mathcal{L}=\partial \mathcal{F}/\partial \theta^{(t)}$.

We now consider as a target system a $\phi^{4}$ lattice action $\mathcal{A}$  defined as 
\begin{equation} \label{eq:fullaction}
\mathcal{A}= \sum_{k=1}^{5} g_{k}\mathcal{A}^{(k)}=  g_{1} \sum_{\langle ij \rangle_{nn} } \phi_{i} \phi_{j} + g_{2} \sum_{i} \phi_{i}^{2}  +  g_{3} \sum_{i} \phi_{i}^{4}  + g_{4} \sum_{\langle ij \rangle_{nnn} } \phi_{i} \phi_{j} + i g_{5} \sum_{i}  \phi_{i}^{2},
\end{equation}
where $nn$ and $nnn$ denote nearest-neighbor and next-nearest neighbor interactions, the $ g_{4}\mathcal{A}^{(4)}$ term is a longer-range interaction, and the action is complex due to the $ g_{5}\mathcal{A}^{(5)}$ term. The coupling constants can have arbitrary values but for this example we consider $g_{1}=g_{4}=-1$, $g_{2}=1.52425$, $g_{3}=0.175$ and $g_{5}=0.15$, see Ref.~\citep{bachtis-qftml}. 

A first proof-of-principle demonstration is to train a $\phi^{4}$ Markov random field with lattice size $L=4$ in each dimension and randomly initialized coupling constants in the action $S$ to approximate the lattice action $\mathcal{A}_{\lbrace 3 \rbrace}$. After training the $\phi^{4}$ Markov random field for $10^{5}$ epochs, the coupling constants have converged to the anticipated values $g_{1}$, $g_{2}$ and $g_{3}$ with precision of order or magnitude $10^{-8}$ (see Ref.~\citep{bachtis-qftml}) and the two systems have therefore become identical.

\begin{figure}[t]
\center
\includegraphics[width=8.6cm]{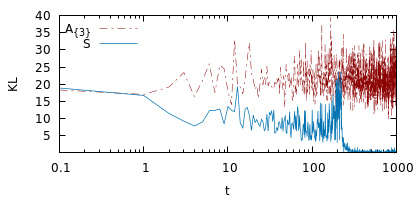}
\caption{\label{fig:kl} Estimated Kullback-Leibler divergence versus epoch $t$ on logarithmic scale. The probability distributions of actions $\mathcal{A}_{\lbrace 3\rbrace }$ and $S$ are compared with the one of $\mathcal{A}_{\lbrace 4\rbrace }$. Only the action $S$ is updated at each epoch based on a finite sample of fixed size. For action $\mathcal{A}_{\lbrace 3\rbrace }$ results are depicted based on a finite sample of equal size to allow for a direct comparison of the two quantities at each epoch $t$. }
\end{figure}

To compare the representational capacity between the $\phi^{4}$ Markov random field with inhomogeneous action $S$ versus a standard $\phi^{4}$ action $\mathcal{A}_{\lbrace 3 \rbrace}$ we can estimate the Kullback-Leibler divergence during the training process. Specifically, we estimate the asymmetric distances of the probability distributions with actions $S$ and $\mathcal{A}_{\lbrace 3 \rbrace}$ against the probability distribution of the action $ \mathcal{A}_{\lbrace 4 \rbrace}$, which includes longer-range interactions. The results are depicted in Fig.~\ref{fig:kl}, where the inhomogeneous action $S$ approximates the target system exceedingly better than the standard action, indicating that it has increased representational capacity.

\begin{figure}[b]
\includegraphics[width=7.5cm]{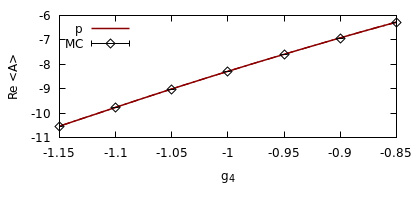}
\includegraphics[width=7.5cm]{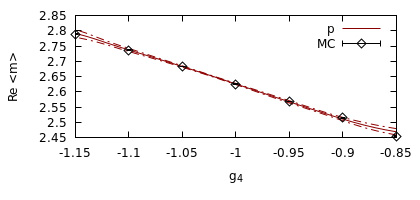}
\caption{\label{fig:re1} Real part of the complex lattice action $\mathcal{A}$ (left) and  real part of the magnetization $m$ (right) versus coupling constant $g_{4}$.  The statistical errors are comparable with the width of the line. }
\end{figure}

We will now investigate how accurately a $\phi^{4}$ Markov random field has approximated a target probability distribution of action $\mathcal{A}_{\lbrace 4\rbrace }$ which includes longer-range interactions. Explicitly, we will investigate if reweighting can be implemented based on the approximating action $S$ to extrapolate in the parameter space of the full action $\mathcal{A}$. We consider the following reweighting procedure:
\begin{equation}\label{eq:rewfull}
\langle O \rangle= \frac{\sum_{l=1}^{N} O_{{l}} \exp[S_{{l}}-g_{j}'\mathcal{A}_{{l}}^{(j)}- \sum_{k=1,k \neq j}^{5}g_{k}\mathcal{A}^{(k)}_{{l}}]}{\sum_{l=1}^{N}  \exp[S_{{l}}-g_{j}'\mathcal{A}_{{l}}^{(j)}- \sum_{k=1,k \neq j}^{5}g_{k}\mathcal{A}_{{l}}^{(k)}]}.
\end{equation}
This form of reweighting acts as a correction step between the probability distributions of the actions $S$ and $ \mathcal{A}_{\lbrace 4 \rbrace}$ and simultaneously extrapolates along the trajectory of the coupling constant $g_j'$ and the complex term $g_{5} \mathcal{A}^{(5)}$. Details can be found in Refs.~\citep{bachtis-qftml,bachtis-addml,bachtis-map,bachtis-ext}.

We consider $j=4$ and extrapolate the observables of the action $\mathcal{A}$ and the magnetization $m$ for values of the coupling constant $g_{4}' \in[-1.15,-0.85]$. The results are depicted in Fig.~\ref{fig:re1} where it is evident that they agree within statistical errors with independent calculations on Monte Carlo datasets supplemented with reweighting to the complex action. The results certify that reweighting to the parameter space of the full action $\mathcal{A}$ is possible based on the approximating action $S$.

The remaining question is how to strictly define the permitted reweighting range during an extrapolation in parameter space based on an approximating action $S$. This can be achieved through the construction of weight functions
\begin{equation}
\mathcal{W}(S)=  \frac{\sum_{\Re [\mathcal{A'}], \Im [\mathcal{A'}]}  h(S,\Re [\mathcal{A'}], \Im [\mathcal{A'}])  \exp[S-\Re [\mathcal{A'}]- i \Im [\mathcal{A'}]]}{\sum_{S,\Re [\mathcal{A'}], \Im [\mathcal{A'}]} h(S,\Re [\mathcal{A'}], \Im [\mathcal{A'}])  \exp[S-\Re [\mathcal{A'}]-i\Im [\mathcal{A'}]]},
\end{equation}
where $\mathcal{A'}=g_{j}'\mathcal{A}^{(j)}+ \sum_{k=1,k \neq j}^{5}g_{k}\mathcal{A}^{(k)}$. The quantity $h(S,\Re [\mathcal{A'}], \Im [\mathcal{A'}])$ is a multi-dimensional histogram of the inhomogeneous action $S$ as well as each action term.

\begin{figure}[t]
\includegraphics[width=7.5cm]{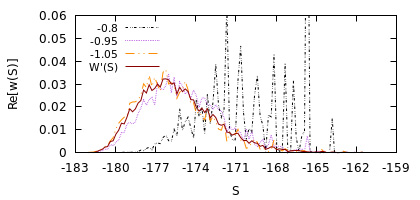}
\includegraphics[width=7.5cm]{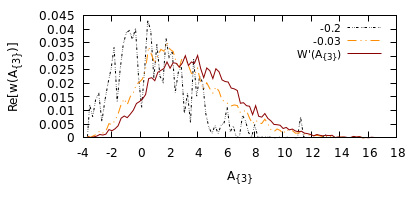}
\caption{\label{fig:hist} Real part of the weight function $\mathcal{W}(S)$ versus lattice action $S$  (left).  Real part of the weight function $\mathcal{W}(\mathcal{A}_{\lbrace 3 \rbrace})$ versus lattice action $\mathcal{A}_{\lbrace 3 \rbrace}$ (right). The results are for considered coupling constants $g_{4}'$. }
\end{figure}

The calculations of the weight functions $\mathcal{W}(S)$ are depicted in Fig.~\ref{fig:hist} (left). We observe that weight functions which are constructed for coupling constants adjacent to $g_{4}=-1$ overlap. In addition, inconsistencies can be observed for $g_{4}'=-0.8$, indicating that reweighting extrapolations beyond the range of this specific value are inacurrate. We remark that histogram reweighting from the action $\mathcal{A}_{3}$ to the full action is impossible due to an insufficient overlap of statistical ensembles. The result is depicted in Fig.~\ref{fig:hist} (right). Inconsistencies in reweighting begin to emerge for the value $g_{4}'=-0.2$, therefore verifying that the parameter space which corresponds to the coupling constant $g_{4}'=-1.0$ in the action   $\mathcal{A}$ cannot be reached from the parameter space of the action $\mathcal{A}_{3}$.

\subsection{Learning with predefined data}

In the previous examples, established based on the minimization of the Kullback-Leibler divergence of Eq.~\ref{eq:kl}, we considered that the probability distribution $q(\phi)$ and the corresponding action $\mathcal{A}$ were known. However, there exist cases where one might have already obtained a dataset, for instance through a Monte Carlo simulation, as experimental data or based on a set of images. These datasets then encode an empirical probability distribution $q(\phi)$ whose form is unknown. One can then ask if we can learn the appropriate coupling constants in the action $S$ of the $\phi^{4}$ theory that are able to approximate this target probability distribution $q(\phi)$ which is explicitly encoded in the available dataset. The problem can be solved using an alternative expression of the Kullback-Leibler divergence:
\begin{equation} \label{eq:klopp}
KL(q || p) = \int_{-\infty}^{\infty} {q(\bm{\phi})} \ln \frac{q(\bm{\phi})}{p(\bm{\phi}; \theta)} d\bm{\phi}  \geq 0.
\end{equation}

By substituting and taking the derivative in terms of the variational parameters $\theta$ we obtain:
\begin{equation}
\frac{\partial \ln p(\phi ; \theta)}{\partial \theta} = \Big\langle \frac{\partial S}{\partial \theta} \Big\rangle_{p(\phi ; \theta)} -  \frac{\partial S}{\partial \theta},
\end{equation}
and the update rule of the parameters $\theta$  at each epoch $t$ is given according to Eq.~\ref{eq:gas}, where $\mathcal{L}=-\partial \ln p(\phi ; \theta^{(t)})/{\partial \theta^{(t)}}$.

\begin{figure}[t]
\includegraphics[width=14.2cm]{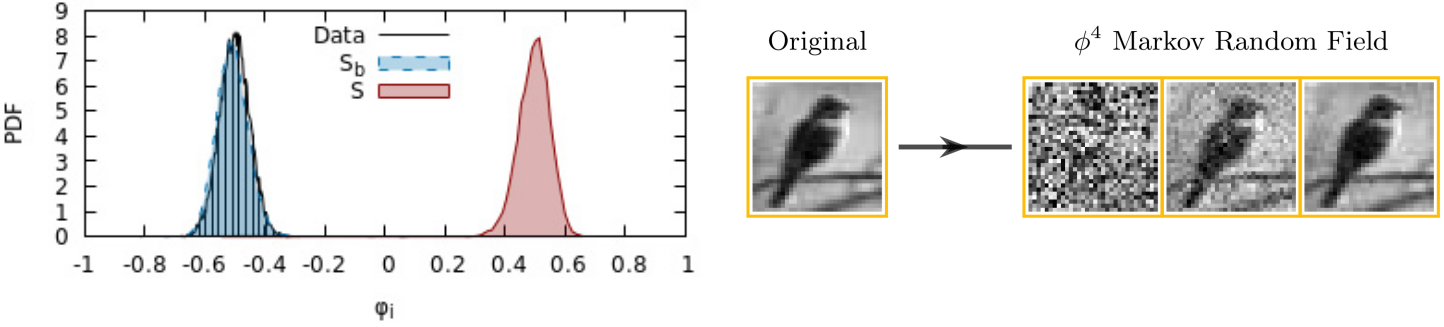}
\caption{\label{fig:2} Probability density function versus lattice value $\phi_{i}$ for a Euclidean action $S$ that is $Z_{2}$ invariant and $S_{b}$ which includes a local symmetry-breaking term (left). Original image and equilibration of the Markov random field after $1$, $10$ and $50$ steps (right).}
\end{figure}

We reiterate that the method is applicable to an arbitrary probability distribution $q(\phi)$. For instance, one can consider as $q(\phi)$ the empirical distribution encoded in a dataset sampled from a Gaussian distribution with $\mu=-0.5$ and $\sigma=0.05$.  We anticipate, due to the $Z_{2}$ invariance in the lattice action, that the symmetric probability distribution is equiprobable in being reproduced. To explicitly reproduce the desired distribution a term of the form $\sum_{i} r_{i} \phi_{i}$ can be included in the action: this term breaks the symmetry of the system explicitly. The results are depicted in Fig.~\ref{fig:2} (left) where we observe the anticipated behaviour.

 We additionally demonstrate the applicability of the method in an image from the CIFAR-10 dataset. Specifically, we search for the optimal values of the coupling constants in the $\phi^{4}$ action that are able to reproduce the image as a configuration in the equilibrium distribution. The results are verified in Fig.~\ref{fig:2} (right). We remark the method can be potentially extended to applications pertinent to the renormalization group~\citep{bachtis-irg}, where it can be used to calculate the appropriate coupling constants which describe renormalized systems.

\subsection{Machine learning with $\phi^{4}$ neural networks}

To derive a neural network architecture from the $\phi^{4}$ theory we will introduce within the graph structure a new set of variables $h_{j}$ which are called hidden. In addition, we will consider that the topology of the graph structure is that of a bipartite graph, and that the interactions are exclusively between the visible $\phi$ and the hidden $h$ variables. The lattice action which describes this system is
\begin{equation}\label{eq:many}
S (\phi,h ; \theta) =  -\sum_{i,j} w_{ij} \phi_{i}h_{j}  + \sum_{i} r_{i} \phi_{i}  + \sum_{i} a_{i} \phi_{i}^{2} + \sum_{i} b_{i} \phi_{i}^{4}  + \sum_{j} s_{j} h_{j} + \sum_{j} m_{j} h_{j}^{2} + \sum_{j} n_{j} h_{j}^{4}.
\end{equation}

By setting certain coupling constants to zero or by considering that the visible or the hidden variables are binary one can arrive at different neural network architectures. These architectures have been extensively studied in computer science, for instance, see Ref.~\citep{fischer-pgm} and references therein. Specifically, if $b_i=n_j=0$ one obtains a Gaussian-Gaussian restricted Boltzmann machine and if $b_i=n_j=m_j=0$ and $h_{j} \in \lbrace-1,1\rbrace$ then the architecture is a Gaussian-Bernoulli restricted Boltzmann machine. We emphasize that the $\phi^{4}$ theory reduces to an Ising model under the limit $\kappa_{L}$ positive and fixed, $\lambda_{L} \rightarrow \infty$ and $\mu_{L}^{2} \rightarrow - \infty$. This equivalence could potentially shed novel insights into this class of machine learning algorithms which were initially inspired by Ising models.

\begin{figure}[t]
\center
\includegraphics[width=8.6cm]{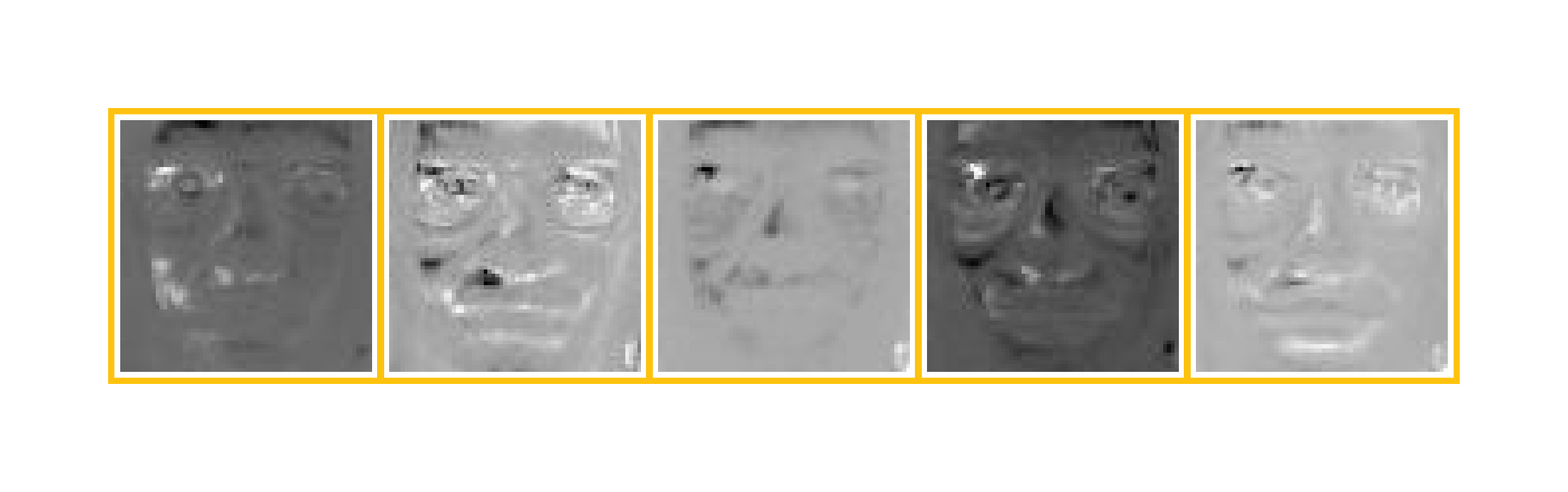}
\caption{\label{fig:faces} Example features learned in the hidden layer of the $\phi^{4}$ neural network.}
\end{figure}

To demonstrate that the $\phi^{4}$ neural network of Eq.~\ref{eq:many} is capable of extracting meaningful features during the learning of a probability distribution with predefined data, we train it on the first forty examples of the Olivetti faces dataset. A representative example of the learned features, i.e. the coupling constants $w_{ij}$ for a fixed $j$, are depicted in Fig.~\ref{fig:faces}. We observe that the neural network has learned features which correspond to abstract face shapes.

\section{Conclusions}

In this contribution we derived machine learning algorithms and neural networks from lattice field theories, opening up the opportunity to investigate machine learning directly within quantum field theory~\citep{bachtis-qftml}. Through the Hammersley-Clifford theorem, we established a precise equivalence between the $\phi^{4}$ theory on a square lattice and the mathematical framework of Markov random fields. We additionally derived neural networks  from the $\phi^{4}$ theory which can be viewed as generalizations of a notable class of machine learning algorithms, namely that of restricted Boltzmann machines. Finally, applications pertinent to physics and to computer science were presented.

Numerous research directions can be envisaged. These include the derivation of machine learning algorithms from lattice gauge theories, the investigation of the inhomogeneity in the coupling constants of these systems and the exploration for the emergence of novel phase transitions within quantum field-theoretic machine learning. Finally, through the construction of quantum fields in Minkowski space from Markov fields in Euclidean space, quantum field-theoretic machine learning can be additionally explored based on a mathematical perspective, one that was firstly established within constructive quantum field theory and probability theory.

In summary, the derivation of quantum field-theoretic machine learning algorithms opens up a new research avenue, that of developing a mathematical and computational framework of machine learning within quantum field theory. As a result, the work paves the way for nonperturbative investigations of the dynamics which describe neural networks and machine learning algorithms, a prospect which can be achieved from the perspective of physics and within the highly successful framework of lattice field theory.

\section{\label{sec:level5}Acknowledgements}
The authors received funding from the European Research Council (ERC) under the European Union's Horizon 2020 research and innovation programme under grant agreement No 813942. The work of GA and BL has been supported in part by the UKRI Science and Technology Facilities Council (STFC) Consolidated Grant ST/T000813/1. The work of BL is further supported in part by the Royal Society Wolfson Research Merit Award WM170010 and by the Leverhulme Foundation Research Fellowship RF-2020-461\textbackslash 9. Numerical simulations have been performed on the Swansea SUNBIRD system. This  system is part of the Supercomputing Wales project, which is part-funded by the European Regional Development Fund (ERDF) via Welsh Government.

\end{document}